%% file: main.tex
\documentclass[10pt,twocolumn,letterpaper]{article}

\usepackage[pagenumbers]{cvpr} 
\usepackage{algorithm}
\usepackage{algorithmic}
\usepackage{amsmath}
\usepackage{amssymb}
\usepackage{booktabs}
\usepackage{multirow}


\definecolor{cvprblue}{rgb}{0.21,0.49,0.74}
\usepackage[pagebackref,breaklinks,colorlinks,allcolors=cvprblue]{hyperref}

\def\paperID{*****} 
\def\confName{CVPR}
\def\confYear{2025}

\title{AnomalyMoE: Towards a Language-free Generalist Model \\ for Unified Visual Anomaly Detection}

\author{
Zhaopeng Gu$^{1,2}$~~~~
Bingke Zhu$^{1,2}$~~~~
Guibo Zhu$^{1,2}$~~~~
Yingying Chen$^{1,2}$~~~~\\
Wei Ge$^{1,2}$~~~~
Ming Tang$^{1,2}$~~~~
Jinqiao Wang$^{1,2,3}$\\
  $^{1}$~Foundation Model Research Center, Institute of Automation,\\ Chinese Academy of Sciences, Beijing, China \\ 
  $^{2}$~School of Artificial Intelligence, University of Chinese Academy of Sciences, Beijing, China\\
  $^{3}$~Objecteye Inc., Beijing, China\\
  {\tt\small  \{guzhaopeng2023, gewei2023\}@ia.ac.cn} \\
  {\tt\small \{bingke.zhu, gbzhu, yingying.chen, tangm, jqwang\}@nlpr.ia.ac.cn} \\
}

\begin{document}
\maketitle

\begin{abstract}
Anomaly detection is a critical task across numerous domains and modalities, yet existing methods are often highly specialized, limiting their generalizability. These specialized models, tailored for specific anomaly types like textural defects or logical errors, typically exhibit limited performance when deployed outside their designated contexts. To overcome this limitation, we propose AnomalyMoE, a novel and universal anomaly detection framework based on a Mixture-of-Experts~(MoE) architecture. Our key insight is to decompose the complex anomaly detection problem into three distinct semantic hierarchies: local structural anomalies, component-level semantic anomalies, and global logical anomalies. AnomalyMoE correspondingly employs three dedicated expert networks at the patch, component, and global levels, and is specialized in reconstructing features and identifying deviations at its designated semantic level. This hierarchical design allows a single model to concurrently understand and detect a wide spectrum of anomalies. Furthermore, we introduce an Expert Information Repulsion~(EIR) module to promote expert diversity and an Expert Selection Balancing~(ESB) module to ensure the comprehensive utilization of all experts. Experiments on 8 challenging datasets spanning industrial imaging, 3D point clouds, medical imaging, video surveillance, and logical anomaly detection demonstrate that AnomalyMoE establishes new state-of-the-art performance, significantly outperforming specialized methods in their respective domains.
\end{abstract}

\section{Introduction}

\begin{figure}[t]
  \centering
   \includegraphics[width=0.9\linewidth]{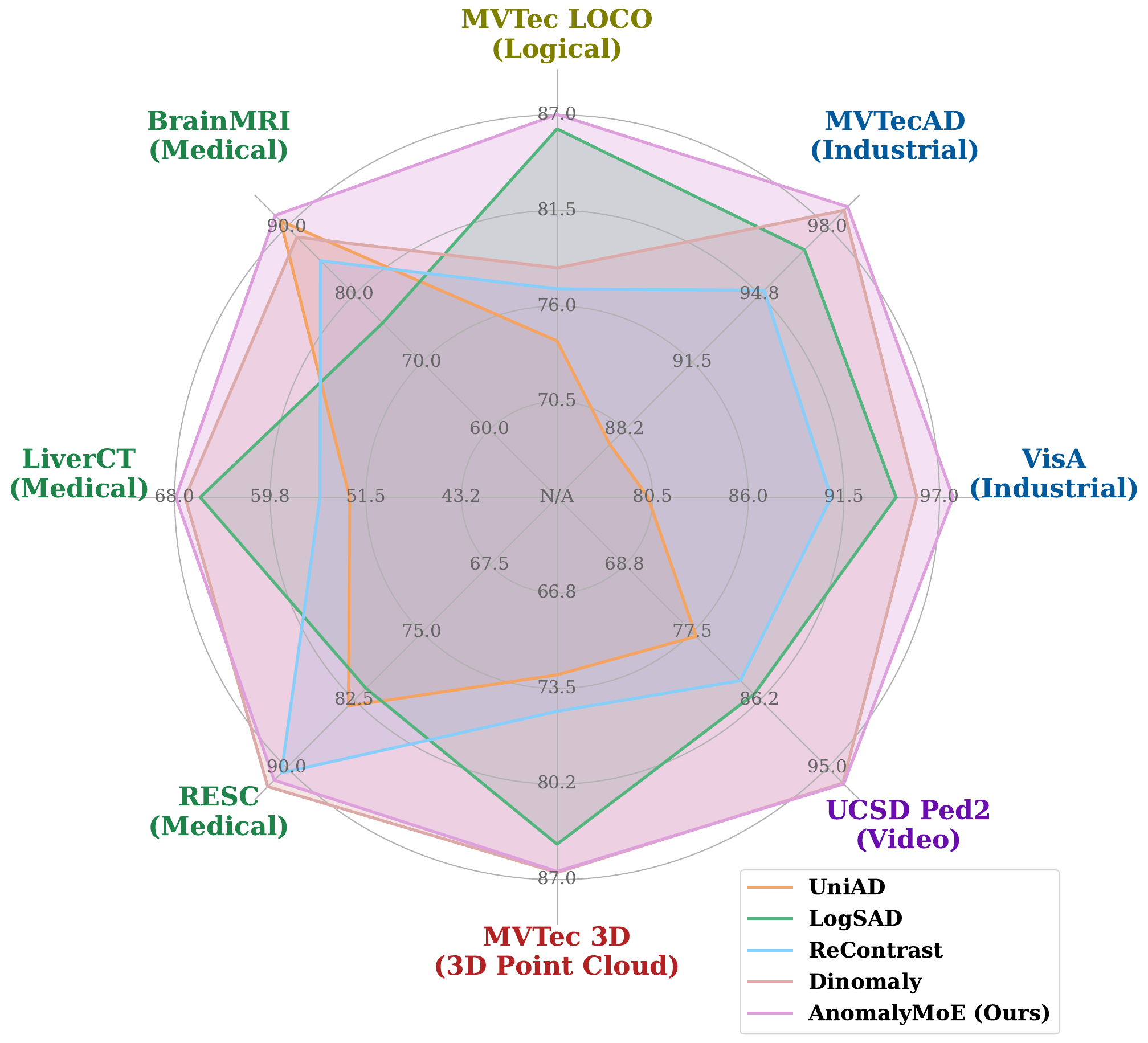}
   \caption{Performance comparison of AnomalyMoE with state-of-the-art anomaly detection methods on a diverse suite of 8 multi-modal, multi-domain datasets.}
   \label{fig:radar}
\end{figure}

\begin{figure*}[t]
  \centering
   \includegraphics[width=\linewidth]{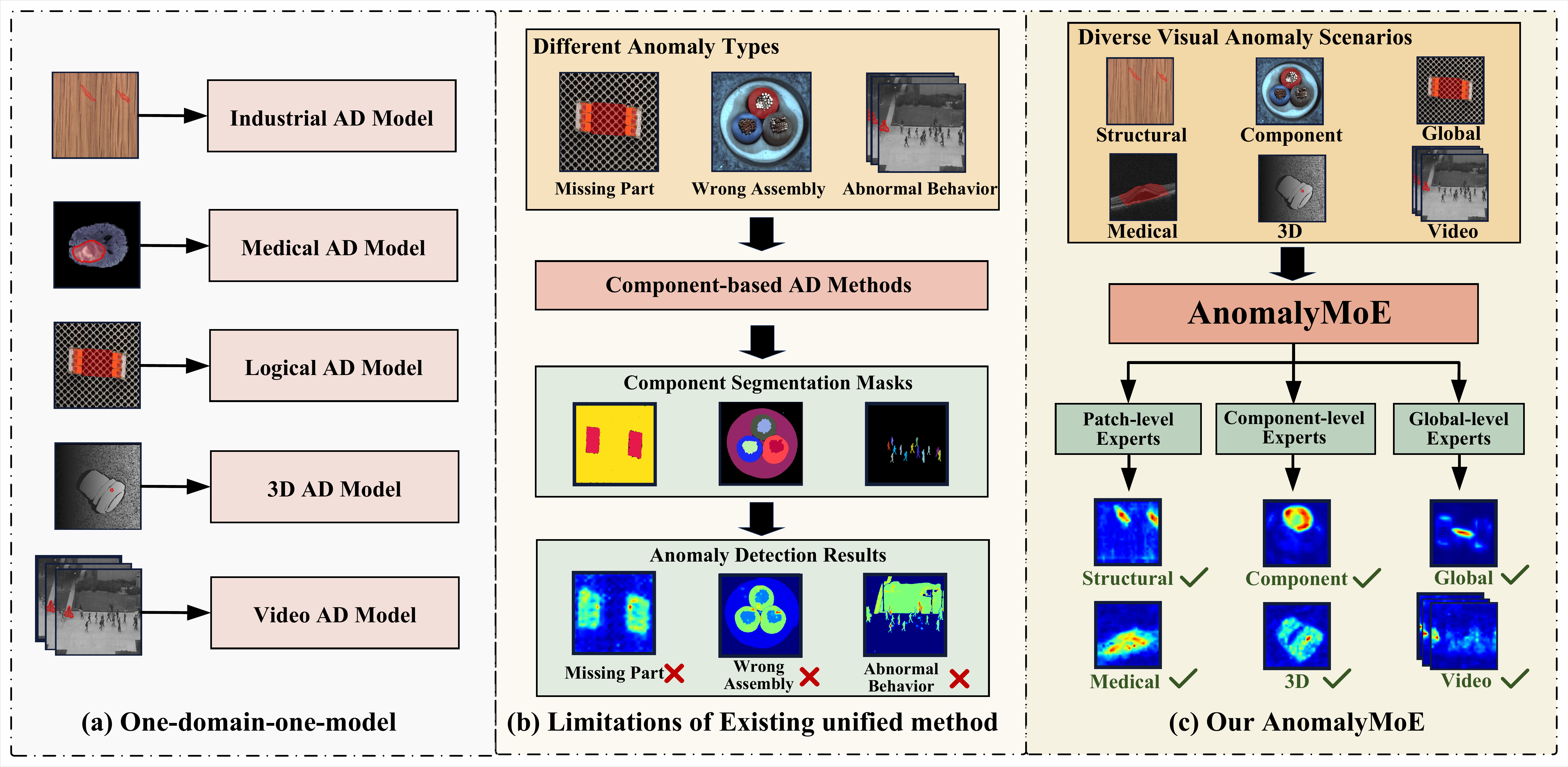}
   \caption{Illustration of the motivation for AnomalyMoE. (a) The conventional one-domain-one-model paradigm, where specialized models are not generalizable. (b) Existing unified methods that typically rely on component-level analysis, failing to detect global or compositional errors (e.g., missing components or incorrect assembly). (c) Our proposed AnomalyMoE, a single, universal model featuring a hierarchical Mixture of Experts. By dedicating experts to patch, component, and global semantic levels, it achieves comprehensive anomaly detection across diverse visual anomaly scenarios.}
   \label{fig:motivation}
\end{figure*}

Anomaly detection aims to identify anomalous samples that deviate from normal patterns. It is widely applied across various fields of production and daily life, including industrial defect detection~\cite{gu2024anomalygpt, zhu2024adformer}, logical anomaly detection~\cite{zhang2025towards, liu2023component}, medical image diagnosis~\cite{bao2024bmad, huang2024adapting}, and video surveillance~\cite{yang2024follow}, etc. Spanning multiple modalities such as images~\cite{gu2025univad}, videos~\cite{yang2024follow}, and 3D point clouds~\cite{ye2025po3ad}, it represents a comprehensive, cross-domain, and cross-modal task with significant practical value, making it a prominent research topic in both academia and industry.

As illustrated in Figure~\ref{fig:motivation}(a), existing anomaly detection methods are often highly specialized for particular domains and modalities~\cite{zhu2024adformer, gu2024filo}. This specialization, while effective for specific tasks, inherently limits their generalizability. To address this, recent efforts have explored unified anomaly detection frameworks. However, these approaches still exhibit critical limitations. Some are confined to patch-level reconstruction, effectively capturing structural flaws but failing to comprehend higher-level logical anomalies. Conversely, others that excel at logical anomaly detection, such as UniVAD~\cite{gu2025univad} and LogSAD~\cite{zhang2025towards}, heavily rely on component segmentation, rendering them incapable of identifying issues like missing components or abnormal arrangements, as shown in Figure~\ref{fig:motivation}(b). Furthermore, these advanced models increasingly depend on large vision-language or language models to interpret component semantics, introducing substantial computational overhead and a reliance on non-visual priors. This complexity poses a barrier to deployment and may not be optimal for purely visual tasks.

To address these issues, we propose AnomalyMoE, a unified and language-free framework that reconceptualizes the Mixture-of-Experts (MoE) architecture for anomaly detection. Its core idea is to decompose the task into a hierarchy of semantic sub-problems. As illustrated in Figure~\ref{fig:motivation}(c), this allows a single model with dedicated experts for local, component, and global semantics to comprehensively handle diverse anomaly types, directly addressing the shortcomings of prior work with superior performance and efficiency.

Specifically, we first categorize ``anomalies" into three semantic levels: 1) Local structural anomalies, which manifest as fine-grained textural or pixel-level deviations (e.g., surface defects); 2) Component-level semantic anomalies, which appear as errors in an object part or region (e.g., incorrect components on a circuit board); and 3) Global logical anomalies, characterized by errors in global semantics (e.g., the incorrect assembly of correct components or abnormal behaviors in video surveillance). To detect these levels, AnomalyMoE employs three corresponding expert networks operating on a feature reconstruction paradigm. A Transformer-based~\cite{vaswani2017attention} decoder expert targets fine-grained local details, while two autoencoder-based~\cite{minhas2020semi} experts learn holistic representations for component and global semantics. Trained exclusively on normal samples, AnomalyMoE identifies anomalies via reconstruction errors. To ensure the experts are both diverse and fully utilized, we further introduce an Expert Information Repulsion (EIR) module to maximize their specialization and an Expert Selection Balancing (ESB) module to prevent model collapse. This integrated design not only enables targeted detection at each semantic level but also allows their features to mutually enhance one another, leading to robust and superior performance.

We conduct experiments on 8 datasets across diverse domains, including industrial imaging~\cite{bergmann2021mvtec}, industrial 3D~\cite{bergmann2021mvtec3d}, medical imaging~\cite{bao2024bmad}, video surveillance~\cite{wang2010anomaly}, and logical anomaly detection~\cite{bergmann2022beyond}. The results demonstrate that AnomalyMoE achieves top performance in anomaly detection across multiple domains. Furthermore, the experts at different semantic levels successfully detect their corresponding types of anomalies, showcasing the comprehensive capability of AnomalyMoE.

Our contributions can be summarized as follows:
\begin{itemize}
    \item We propose AnomalyMoE, a language-free, generalist framework for visual anomaly detection, significantly advancing the field of unified anomaly detection with its MoE architecture.
    \item AnomalyMoE's three-level experts interpret test samples from different semantic levels to accurately detect various types of anomalies. The Expert Information Repulsion and Expert Selection Balancing modules further enhance the diversity and balance among the experts. 
    \item Extensive experiments on 8 datasets across domains such as industrial image, industrial 3D, medical image, video surveillance, and logical AD demonstrate that AnomalyMoE surpasses specialized methods in each respective field, achieving state-of-the-art performance.
\end{itemize}

\section{Related Work}

\subsection{Anomaly Detection}

Anomaly detection is a comprehensive task spanning multiple modalities and domains~\cite{bergmann2021mvtec, bergmann2021mvtec3d, wang2010anomaly}. However, the nature of anomalies, ranging from industrial defects and logical errors to medical pathologies and abnormal behaviors, varies drastically in features and semantic levels. This diversity has led to a fragmented research landscape following a ``one domain, one method" paradigm. For instance, industrial vision methods focus on patch-level feature analysis~\cite{gu2024filo, gu2025filo++, defard2021padim, wang2025cnc} for local structural defects, while logical anomaly detection relies on component-based segmentation~\cite{liu2023component, hsieh2024csad, kim2024few} to identify relational errors. Similarly, video surveillance uses sequence models like RNNs to capture temporal deviations~\cite{naji2022spatio}, and other domains like medical imaging or 3D point clouds employ specialized techniques targeting semantic or geometric properties~\cite{liang2025look, zhou2024pointad}. This specialization has created a landscape of incompatible specialized tools, hindering the development of universal anomaly detection.

\begin{figure*}[t]
  \centering
   \includegraphics[width=\linewidth]{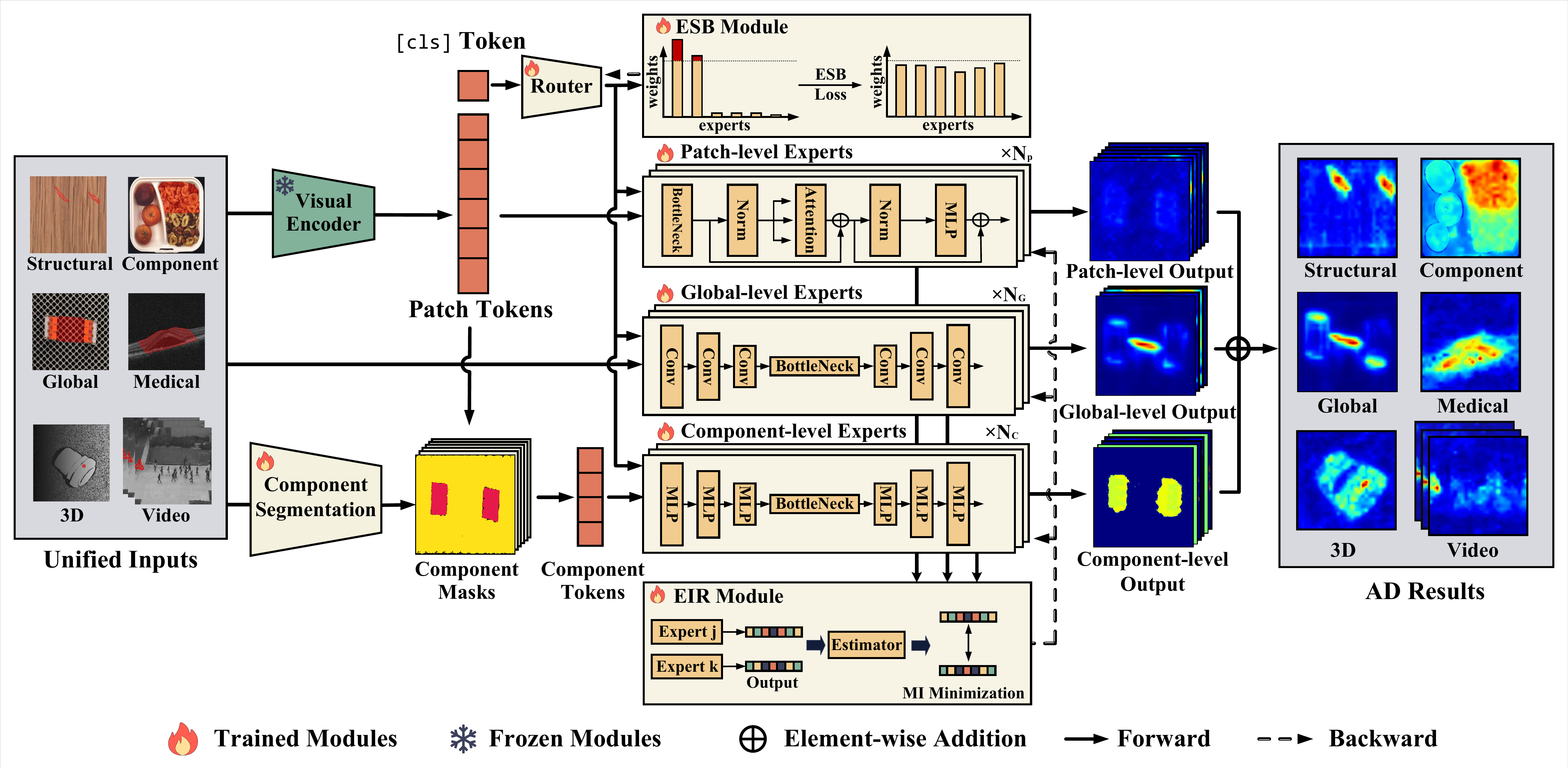}
   \caption{The overall architecture of AnomalyMoE. An input sample is processed by a frozen encoder to extract patch and \texttt{[cls]} embeddings. The \texttt{[cls]} token guides a router to activate a subset of three heterogeneous expert groups (Patch, Component, and Global level). Each expert produces an anomaly map, which are aggregated for the final result. The training is regularized by the Expert Selection Balancing (ESB) and Expert Information Repulsion (EIR) modules.}
   \label{fig:arch}
\end{figure*}

The pursuit of ``a single model for all anomalies" is a key research frontier. UniAD~\cite{you2022unified} is an early pioneer, using a Transformer model with learnable queries to address the ``identity mapping" problem for multi-class detection. ReContrast~\cite{guo2023recontrast} and UniNet~\cite{wei2025uninet} integrate contrastive learning and domain adaptation, while Dinomaly~\cite{guo2025dinomaly} proposes a simpler reconstruction network. However, these methods, being confined to patch-level reconstruction, primarily excel at industrial defects. To bridge this gap, UniVAD~\cite{gu2025univad} and LogSAD~\cite{zhang2025towards} augment their frameworks with a component-level branch. Specifically, these methods leverage large vision-language models to interpret component semantics. Despite this advance, they face critical issues: a heavy reliance on segmentation accuracy, which fails on ``unsegmentable" anomalies, and a lack of a global perspective to assess overall assembly. Moreover, their dependence on large language models introduces significant computational overhead. Our efficient, language-free approach overcomes these limitations while delivering superior performance.

In summary, existing unified models are either limited to local details or dependent on flawed, non-global analysis. They fail to provide a framework covering the three hierarchical levels of local structure, component semantics, and global logic. This research gap is the core motivation for our proposed AnomalyMoE.

\subsection{Mixture-of-Experts}
The Mixture-of-Experts (MoE) is a conditional computation architecture featuring a gating network and multiple expert networks~\cite{shazeer2017outrageously}. By sparsely activating experts based on the input, MoE enables massive model scaling with constant computational cost. GShard~\cite{lepikhin2020gshard} first combines MoE with the Transformer~\cite{vaswani2017attention} architecture, enhancing multilingual machine translation by adding multiple parallel FFNs. Switch Transformer~\cite{fedus2022switch} integrates an MoE architecture into the T5~\cite{raffel2020exploring} model, yielding a more powerful pretrained language model. V-MoE~\cite{riquelme2021scaling} brings this paradigm to the vision domain, leading to the training of the largest vision models to date. The first application in anomaly detection, Adapted-MoE~\cite{lei2024adapted}, uses homogeneous experts to handle intra-class variations within datasets.

While existing MoE applications focus on model scaling, we reframe its purpose for effective problem decomposition. The value of MoE in our framework stems not from an increase in parameter count, but from its ability to dissect the complex anomaly detection task into distinct, semantically clear sub-problems. We assign specialized, heterogeneous experts to each sub-task and introduce Expert Information Repulsion and Expert Selection Balancing modules to govern their collaboration. This approach re-purposes the MoE architecture, transforming it from a tool for achieving greater model capacity into a sophisticated framework that performs hierarchical and multi-level semantic analysis.

\section{Method}
\label{sec:method}

As illustrated in Figure~\ref{fig:arch}, the core of AnomalyMoE is a hierarchical Mixture-of-Experts architecture. An input sample is first processed by a frozen, pre-trained visual encoder to extract patch embeddings and a global \texttt{[cls]} embedding. The \texttt{[cls]} embedding then directs a trainable router to orchestrate a parallel ensemble of three heterogeneous expert groups, each tailored to a distinct semantic hierarchy: local structure, component semantics, and global logic. The training process is optimized by two novel auxiliary modules, Expert Information Repulsion (EIR) and Expert Selection Balancing (ESB), which ensure functional diversity and balanced utilization. We will first detail this novel mechanism, followed by the specific implementations of the expert groups, and conclude with the final training objective.

\subsection{Expert Routing and Collaboration Mechanism}
To effectively manage expert networks and encourage them to learn diverse representations, we design a sophisticated expert routing and collaboration mechanism. At its core is a dynamic routing module, supplemented by the Expert Selection Balancing module and the Expert Information Repulsion module. These components work in synergy to address common challenges in MoE architectures, such as expert functional redundancy and training instability.

\noindent\textbf{Top-K Sparse Expert Routing. }
The Router module is a lightweight, trainable feed-forward network that dynamically allocates weights to experts based on the global \texttt{[cls]} embedding, denoted as $E_{\text{[cls]}}$. To ensure efficiency, we employ a Top-K sparse routing mechanism. The router first generates a vector of scores, or \texttt{logits}, for all experts by applying a learnable weight matrix $\mathbf{W}_{\text{gate}}$. We then activate only the $K$ experts with the highest logit values (where K=3 in our implementation). The final gating weights are computed by applying a Softmax function exclusively to the logits of these selected experts. This process is formalized as:
\begin{equation}
\text{Indices}_{\text{topK}} = \text{TopK}(\mathbf{W}_{\text{gate}} \cdot E_{\text{[cls]}}, K),
\end{equation}
\begin{equation}
G_j = 
\begin{cases} 
\frac{\exp(\text{logit}_j)}{\sum_{k \in \text{Indices}_{\text{topK}}} \exp(\text{logit}_k)}, & \text{if } j \in \text{Indices}_{\text{topK}}, \\
0, & \text{otherwise},
\end{cases}
\end{equation}
where $\text{logit}_j$ is the score for the $j$-th expert, $\text{Indices}_{\text{topK}}$ contains the indices of the selected experts, and $G_j$ is the final gating weight for the $j$-th expert. By activating only a small subset of experts per input, this approach significantly reduces both computational load and memory usage.

\noindent\textbf{Expert Selection Balancing. }
A common challenge in MoE training is router convergence to a state where only a few experts are consistently activated. To prevent this, our Expert Selection Balancing (ESB) module introduces a multi-part load balancing auxiliary loss, $\mathcal{L}_{\text{ESB}}$. For a training batch of size $B$, we define:
1) Expert Importance $P_j$: The average gating probability for the $j$-th expert, $P_j = \frac{1}{B}\sum_{b=1}^{B} G_{b,j}$.
2) Expert Count $C_j$: The number of samples assigned to the $j$-th expert.
The total ESB loss is a weighted sum:
\begin{equation}
    \label{eq:esb_loss}
    \mathcal{L}_{\text{ESB}} = \lambda_{\text{imp}} \mathcal{L}_{\text{importance}} + \lambda_{\text{load}} \mathcal{L}_{\text{load}} + \lambda_{z} \mathcal{L}_{z}.
\end{equation}
Here, the importance loss, $\mathcal{L}_{\text{importance}} = N_{\text{exp}} \sum_j P_j^2$, encourages a uniform distribution of expert probabilities, where $N_{\text{exp}}$ is the total number of experts. The load loss, $\mathcal{L}_{\text{load}} = \sum_j \max(C_j - \text{Capacity}, 0)^2$, enforces a hard capacity constraint by penalizing any expert whose count $C_j$ exceeds a defined $Capacity$. Finally, the z-loss, $\mathcal{L}_{z} = \frac{1}{B} \sum_{b,j} (\text{logit}_{b,j})^2$, is a regularization term applied to the pre-softmax logits to enhance training stability. The terms $\lambda_{\text{imp}}$, $\lambda_{\text{load}}$, and $\lambda_{z}$ are set to 1 in our experiments.

\noindent\textbf{Expert Information Repulsion. }
While ESB ensures expert utilization, the EIR module promotes their functional specialization by reducing redundancy. Inspired by information-theoretic methods~\cite{zhang2024multiple}, EIR minimizes the Mutual Information (MI) between the output representations of different experts within the same group. Let $Z_j$ and $Z_k$ be random variables representing the output feature representations of the $j$-th and $k$-th experts, respectively. Our goal is to minimize $I(Z_j; Z_k)$.

Since direct MI computation is intractable, we employ the Contrastive Log-ratio Upper Bound (CLUB)~\cite{cheng2020club} to estimate and minimize this value. This is achieved using a variational network, $q_{\theta}(z_k | z_j)$, to approximate the conditional probability, where $z_j$ and $z_k$ are specific samples of the random variables $Z_j$ and $Z_k$. The MI upper bound is then calculated via Monte Carlo estimation:
\begin{equation}
\label{eq:club_loss}
\begin{split}
    I_{\text{CLUB}}(Z_j; Z_k) \approx{} & \mathbb{E}_{p(z_j, z_k)}[\log q_{\theta}(z_k | z_j)] \\
    & - \mathbb{E}_{p(z_j)p(z_k)}[\log q_{\theta}(z_k | z_j)].
\end{split}
\end{equation}
The first term is the expected log-likelihood over matched sample pairs $(z_j^{(b)}, z_k^{(b)})$ from the batch, and the second is over non-matched pairs created by permutation. The total EIR loss, $\mathcal{L}_{\text{EIR}}$, is the sum of these CLUB estimates over all expert pairs within each group. By minimizing this loss, we push the representation spaces of experts apart, compelling them to learn distinct, complementary functions. Further details about the CLUB estimates are provided in Appendix~A.

\subsection{Hierarchical Expert Implementations}
The effectiveness of our framework relies on three groups of heterogeneous experts, each designed to target a specific semantic level of anomalies.

\noindent\textbf{Patch-level Experts. }
This group of $N_p$ experts identifies fine-grained structural anomalies. Their objective is to reconstruct a target feature map $F_{\text{target}} \in \mathbb{R}^{N \times D}$, which fuses multi-level information from the encoder. A key challenge is the ``identity mapping'' problem, where a powerful network can learn to perfectly copy its input, rendering it unable to detect anomalies. To address this, we first corrupt $F_{\text{target}}$ with sequential Gaussian noise $\mathcal{N}(\cdot)$ and dropout $\mathcal{D}(\cdot;p)$ with a dropout rate of $p$, which is set to 0.2 in our experiments:
\begin{equation}
    F_{\text{rec\_in}} = \mathcal{D}(\mathcal{N}(F_{\text{target}}); p).
\end{equation}
Second, within each Transformer-based expert, we replace standard Softmax Attention with Linear Attention~\cite{han2023flatten, guo2025dinomaly}. Conventional Softmax Attention can learn highly localized mappings where a query token attends sharply to its corresponding key, facilitating trivial copying. In contrast, Linear Attention promotes a more diffuse attention distribution by replacing the exponential function with a linear operator. This property compels the model to integrate global context for reconstruction, serving as an effective regularizer while also reducing computational complexity from $O(N^2d)$ to $O(Nd^2)$. The training objective for each patch expert is to minimize the reconstruction loss $\mathcal{L}_p$, defined as the mean cosine distance between the target feature vectors and their reconstructions $F_{\text{rec\_out}}$:
\begin{equation}
    \mathcal{L}_p = \mathbb{E} \left[ \frac{1}{N} \sum_{i=1}^{N} \left( 1 - \frac{F_{\text{target}}(i) \cdot F_{\text{rec\_out}}(i)}{\|F_{\text{target}}(i)\| \|F_{\text{rec\_out}}(i)\|} \right) \right],
\end{equation}
where $i$ is the patch index. During inference, the pointwise calculation of distance yields the anomaly score map $S_p$.

\begin{table*}[]
\centering
\small
\caption{Quantitative comparison of AnomalyMoE with state-of-the-art methods across eight diverse datasets under the unified, single-model setting. We report Image-level and Pixel-level AUC (\%). The best-performing method is highlighted in \textbf{bold}. }
\label{tab:main-results}
\begin{tabular}{@{}cccccccccc@{}}
\toprule
Metric & Dataset & Task & UniAD & Recontrast & UniNet & Dinomaly & UniVAD & LogSAD & \textbf{AnomalyMoE} \\ \midrule
\multirow{8}{*}{\begin{tabular}[c]{@{}c@{}}Image-level\\ (AUC)\end{tabular}} 
 & MVTec-AD   & Industrial & 87.5 & 95.0 & 84.7 & 98.8          & 97.8 & 96.9 & \textbf{99.5} \\
 & VisA       & Industrial & 80.2 & 90.1 & 87.5 & 95.7          & 93.5 & 94.5 & \textbf{98.1} \\
 & MVTec LOCO & Logical    & 71.3 & 79.1 & 77.8 & 78.2          & 71.0 & 86.2 & \textbf{87.5} \\
 & BrainMRI   & Medical    & 90.9 & 85.0 & 54.1 & 88.5          & 80.2 & 75.8 & \textbf{92.1} \\
 & LiverCT    & Medical    & 52.9 & 55.5 & 47.5 & 67.1          & 70.0 & 65.8 & \textbf{71.1} \\
 & RESC       & Medical      & 83.2 & 94.6 & 70.3 & \textbf{92.2} & 85.5 & 81.2 & \textbf{92.2} \\
 & MVTec 3D   & 3D         & 72.5 & 75.1 & 70.0 & 86.5          & 80.2 & 84.5 & \textbf{87.2} \\
 & Ped2       & Video      & 72.5 & 83.7 & 61.7 & \textbf{97.1} & 94.3 & 85.5 & \textbf{97.1} \\ \midrule
\multirow{8}{*}{\begin{tabular}[c]{@{}c@{}}Pixel-level\\ (AUC)\end{tabular}} 
 & MVTec-AD   & Industrial & 94.2 & 96.3 & 91.2 & 97.6          & 96.5 & 97.0 & \textbf{97.7} \\
 & VisA       & Industrial & 96.3 & 90.8 & 87.3 & 96.4          & 98.2 & 97.1 & \textbf{99.0} \\
 & MVTec LOCO & Logical    & 76.4 & 74.1 & 73.2 & 71.5          & 75.1 & 79.4 & \textbf{82.2} \\
 & BrainMRI   & Medical    & 98.0 & 96.3 & 86.4 & 95.8          & 96.8 & 95.5 & \textbf{97.3} \\
 & LiverCT    & Medical    & 96.3 & 96.6 & 95.0 & 97.2          & 96.3 & 96.6 & \textbf{97.2} \\
 & RESC       & Medical      & 95.9 & 94.6 & 78.4 & 95.3          & 94.9 & 91.9 & \textbf{96.0} \\
 & MVTec 3D   & 3D         & 95.3 & 75.1 & 95.2 & 98.7          & 94.6 & 98.2 & \textbf{98.9} \\
 & Ped2       & Video      & 95.4 & 97.1 & 96.3 & 98.3          & 97.4 & 98.3 & \textbf{98.4} \\ \bottomrule
\end{tabular}
\end{table*}

\noindent\textbf{Component-level Experts. }
This group of $N_c$ experts identifies semantic anomalies, where a component's local texture may be normal but its identity is incorrect. To extract features for individual components, we first establish a ``component knowledge base'' for each object class by applying K-Means clustering to the patch embeddings of all its normal training samples. During inference, each patch of a test sample is assigned to the nearest cluster center, generating component masks. We then use these masks to perform Masked Average Pooling on the feature map $F_{\text{target}}$, yielding a single embedding vector $c_k \in \mathbb{R}^D$ for each component.

This expert group consists of $N_c$ identically structured experts, each being a lightweight MLP-based Autoencoder. This architecture is well-suited for processing the non-spatial, vectorized representations of the component embeddings $\{c_k\}$. Through its compressed bottleneck layer, each autoencoder is compelled to learn a low-dimensional manifold of the semantic identity of normal components, abstracting away from visual texture. When a semantically anomalous component is input, its embedding $c_k$ deviates from this learned normal manifold, leading to inaccurate reconstruction. 
The training loss $\mathcal{L}_c$ minimizes the mean cosine distance between the original embeddings and their reconstructions $\hat{c}_{k}$:
\begin{equation}
    \mathcal{L}_c = \mathbb{E} \left[ \frac{1}{K_c} \sum_{k=1}^{K_c} \left( 1 - \cos(c_k, \hat{c}_{k}) \right) \right],
\end{equation}
where $K_c$ is the number of components. The individual component scores $S_c^{(k)}$ are calculated using this same cosine distance during inference.

\noindent\textbf{Global-level Experts. }
This group of $N_g$ experts identifies logical anomalies that require a holistic view, such as incorrect component arrangements. Each expert is a convolutional Autoencoder that takes the entire sample $I$ as input and aims to reconstruct the target feature map $F_{\text{target}}$. The architecture comprises a convolutional encoder, $E_g$, which progressively downsamples the sample to capture holistic spatial information, and a corresponding decoder, $D_g$. The information bottleneck between $E_g$ and $D_g$ forces the model to learn a compact, low-dimensional manifold representing the normal global layout and inter-component relations of a scene. An sample with a global anomaly cannot be encoded onto this normal manifold, leading to a high reconstruction error, which is quantified by the squared Euclidean distance:
\begin{equation}
    \mathcal{L}_g = \mathbb{E}_{I \sim \mathcal{D}_{\text{normal}}} \left[ \|F_{\text{target}} - D_g(E_g(I))\|_2^2 \right].
\end{equation}

During inference, the corresponding anomaly score map $S_g$ is generated by computing this same pointwise squared Euclidean distance between the target and its reconstruction.

\subsection{Final Training Objective}

The final training objective of AnomalyMoE, $\mathcal{L}_{\text{total}}$, combines the weighted reconstruction loss from the activated experts with our two auxiliary regularization losses:
\begin{equation}
\mathcal{L}_{\text{total}} = \sum_{j \in \text{Indices}_{\text{topK}}} G_j \cdot \mathcal{L}_{\text{rec}}^{(j)} + \lambda_{\text{ESB}}\mathcal{L}_{\text{ESB}} + \lambda_{\text{EIR}}\mathcal{L}_{\text{EIR}}.
\end{equation}
Here, $\mathcal{L}_{\text{rec}}^{(j)}$ is the reconstruction loss for the $j$-th activated expert, corresponding to one of the losses ($\mathcal{L}_p$, $\mathcal{L}_c$, or $\mathcal{L}_g$) defined in the previous subsections, depending on the expert's type. It is weighted by its gate score $G_j$. Crucially, the auxiliary losses $\mathcal{L}_{\text{ESB}}$ and $\mathcal{L}_{\text{EIR}}$ are computed using the full pre-softmax logit distribution from the router. This ensures that all experts, including inactive ones, receive gradients for load balancing and information repulsion. This integrated objective guides the framework toward a state of high efficiency, diversity, and functional specialization.

\begin{table*}[]
\centering
\small
\caption{Ablation study on the hierarchical expert structure of AnomalyMoE. 
    We report the (Image-level AUC, Pixel-level AUC) in percent for different combinations of expert groups. 
    The full three-level configuration achieves the best performance across all datasets, confirming the contribution of each expert.}
\label{tab:expert-ablation}
\begin{tabular}{@{}cccccccccc@{}}
\toprule
\multicolumn{3}{c}{Expert Levels} & \multicolumn{7}{c}{Datasets} \\ \midrule
Patch                 & Component             & Global                &          MVTec-AD                 &          VisA             &            MVTec LOCO                 &          BrainMRI                 &         RESC              &               MVTec 3D            &         Ped2              \\ \midrule
$\checkmark$          &                       &                       & (98.2, 96.0)              & (97.0, 98.8)          & (78.2, 75.9)                & (91.6, 97.1)              & (91.2, 95.9)          & (85.7, 98.6)              & (96.2, 98.5)          \\
$\checkmark$          & $\checkmark$          &                       & (98.4, 95.1)              & (97.9, 98.9)          & (84.7, 79.4)                & (89.5, 96.4)              & (88.7, 91.2)          & (85.9, 85.8)              & (96.9, 97.4)          \\
$\checkmark$          &                       & $\checkmark$          & (99.1, 96.9)              & (97.7, 98.9)          & (86.8, 81.1)                & (89.2, 97.1)              & (91.5, 95.5)          & (86.3, 97.3)              & (97.0, 98.3)          \\
\textbf{$\checkmark$} & \textbf{$\checkmark$} & \textbf{$\checkmark$} & \textbf{(99.5, 97.7)}     & \textbf{(98.1, 99.0)} & \textbf{(87.5, 82.2)}       & \textbf{(92.1, 97.3)}     & \textbf{(92.2, 96.0)} & \textbf{(87.2, 98.9)}     & \textbf{(97.1, 98.4)} \\ \bottomrule
\end{tabular}
\end{table*}

\begin{table}[]
\centering
\small
\caption{Ablation study on the ESB and EIR auxiliary modules in AnomalyMoE. 
    The table compares the performance of the full model against versions without either the ESB or EIR loss. Results are reported as (Image-level AUC, Pixel-level AUC) in percent.}
\label{tab:esbir-ablation}
\begin{tabular}{@{}cccc@{}}
\toprule
Dataset    & w/o ESB      & w/o EIR      & \textbf{AnomalyMoE}   \\ \midrule
MVTec-AD   & (99.2, 97.2) &   (99.0, 97.1)           & \textbf{(99.5, 97.7)} \\
VisA       & (97.3, 98.8) &   (97.4, 98.9)           & \textbf{(98.1, 99.0)} \\
MVTec LOCO & (84.2, 81.1) &   (85.6, 92.0)           & \textbf{(87.5, 82.2)} \\
BrainMRI   & (91.2, 97.0) &   (91.6, 97.2)           & \textbf{(92.1, 97.3)} \\
LiverCT    & (69.3, 97.1) &   (69.4, 97.1)           & \textbf{(71.1, 97.2)} \\
RESC       & (91.2, 95.3) &   (91.2, 95.7)           & \textbf{(92.2, 96.0)} \\
MVTec 3D   & (85.7, 98.0) &   (85.9, 98.7)           & \textbf{(87.2, 98.9)} \\
Ped2       & (96.0, 97.9) &   (96.6, 98.3)           & \textbf{(97.1, 98.4)} \\ \bottomrule
\end{tabular}
\end{table}

\section{Experiments}

\subsection{Experimental Setup}
\noindent\textbf{Datasets. }
To comprehensively evaluate the generalizability and performance of AnomalyMoE, we conduct experiments on a diverse suite of eight challenging datasets. These benchmarks span multiple domains, including industrial structural anomalies (MVTec AD~\cite{bergmann2021mvtec}, VisA~\cite{zou2022spot}), 3D point clouds (MVTec 3D-AD~\cite{bergmann2021mvtec3d}), logical anomalies (MVTec-LOCO~\cite{bergmann2022beyond}), medical imaging (BrainMRI~\cite{baid2021rsna}, LiverCT~\cite{bilic2023liver}, RESC~\cite{hu2019automated}), and video surveillance (UCSD Ped2~\cite{wang2010anomaly}). This collection provides a robust and varied testbed for our generalist model, covering a wide spectrum of visual anomaly types. A detailed description of each dataset is provided in Appendix~B.

\noindent\textbf{Evaluation Metrics. }
To ensure a fair and comprehensive comparison with prior work, we adopt standard evaluation protocols for anomaly detection. We evaluate performance using the Area Under the Receiver Operating Characteristic curve (AUC). We report both Image-level AUC to assess the model's detection capability and Pixel-level AUC to evaluate its localization accuracy.

\noindent\textbf{Implementation Details.}
We use a pre-trained DINOv2 model (ViT-B/14)~\cite{oquab2023dinov2} as the frozen visual encoder. All input images are resized to 448$\times$448 and then center-cropped to 392$\times$392. For our MoE architecture, we set the number of experts in each group to $N_p=6$, $N_c=6$, and $N_g=6$, with a Top-K routing of $K=3$. The model is trained for 50,000 iterations using the StableAdamW optimizer with a learning rate of $5 \times 10^{-4}$ and a weight decay of $10^{-4}$. We use a batch size of 16 on 4 NVIDIA A6000 GPUs. The hyperparameters for the auxiliary losses are set to $\lambda_{\text{ESB}}=0.01$ and $\lambda_{\text{EIR}}=0.0001$.

\noindent\textbf{Baselines. }
We compare AnomalyMoE against a comprehensive set of state-of-the-art methods to demonstrate its superiority across different paradigms. Our baselines include top-performing generalist patch-level reconstruction-based models like UniAD~\cite{you2022unified}, ReContrast~\cite{guo2023recontrast}, UniNet~\cite{wei2025uninet}, and Dinomaly~\cite{guo2025dinomaly}, and advanced component-based methods that leverage language priors, specifically UniVAD~\cite{gu2025univad} and LogSAD~\cite{zhang2025towards}. For a fair comparison, we reproduce them using their officially released code under our unified evaluation protocol.

\begin{figure}[!h]
  \centering
   \includegraphics[width=\linewidth]{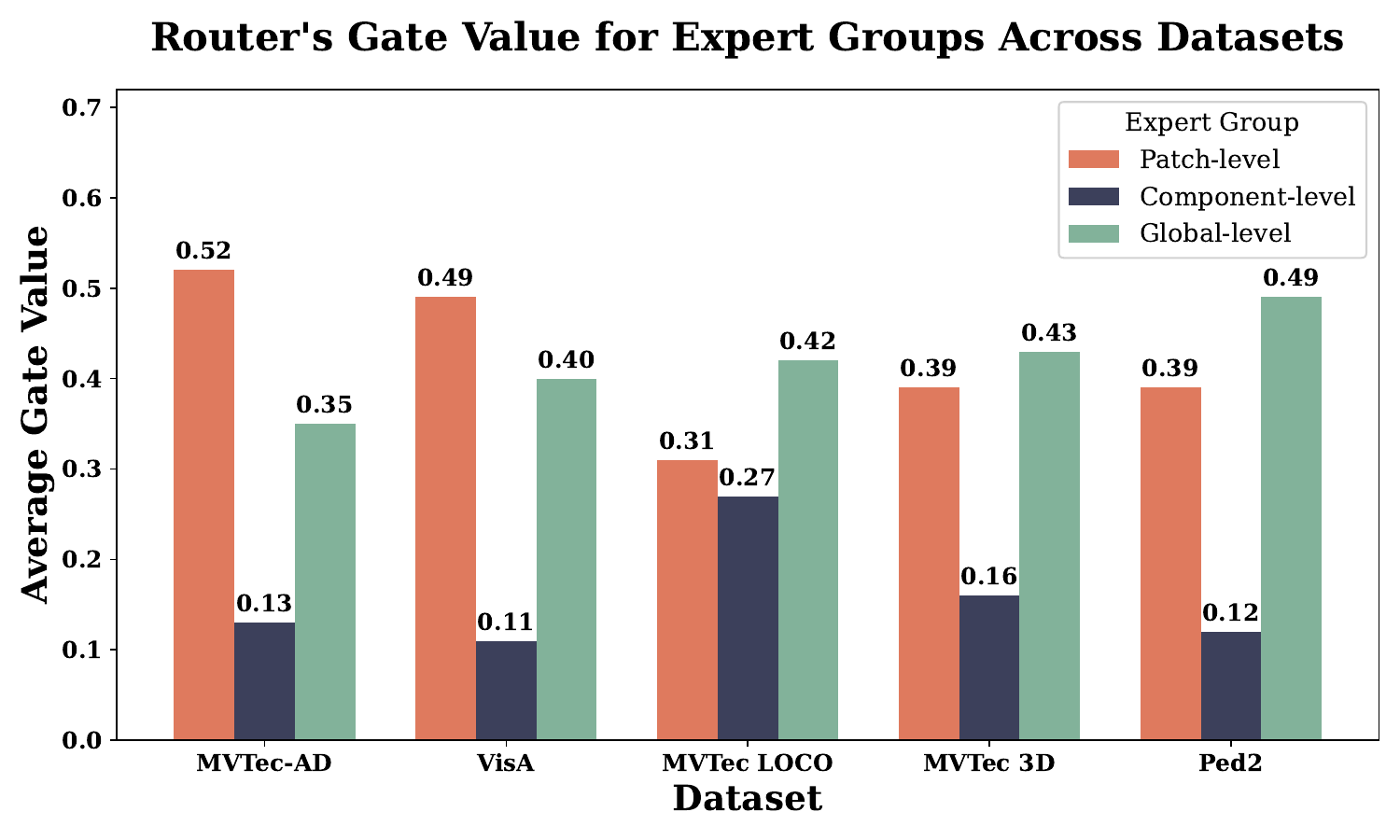}
   \caption{Analysis of the router's behavior. The histogram shows the average gate values assigned to each expert group across various datasets.}
   \label{fig:router_analysis}
\end{figure}

\subsection{Main Quantitative Results}

We present the main quantitative results in Table~\ref{tab:main-results}, establishing AnomalyMoE as the new state-of-the-art in generalist anomaly detection. The limitations of specialized approaches become clear in this unified setting. While reconstruction-based methods like Dinomaly excel on structural datasets, they falter on logical ones. On the other hand, component-based methods like LogSAD performs exceptionally well on its target MVTec-LOCO dataset, yet its performance significantly degrades on other domains. This demonstrates that while specialized methods often suffer a significant performance drop when generalized, AnomalyMoE's hierarchical MoE architecture delivers robust, top-tier performance across all anomaly types, validating its superior design.

Beyond its superior accuracy, AnomalyMoE offers a significant advantage in efficiency. Unlike methods such as UniVAD and LogSAD that depend on computationally expensive large language models, our framework is entirely language-free. This results in a faster inference speed, as detailed in Appendix~C, enhancing its practicality for real-world deployment.

\subsection{Ablation Studies}
To validate our key design choices, we conduct a series of comprehensive ablation studies, with results summarized in Table~\ref{tab:expert-ablation} and Table~\ref{tab:esbir-ablation}. The results first demonstrate the necessity of our hierarchical expert structure. As shown in Table~\ref{tab:expert-ablation}, no single expert level is sufficient for robust, generalist performance. For instance, the Patch-only model excels on the structural anomalies of MVTec AD but fails on the logical tasks in MVTec LOCO. The full three-level configuration consistently outperforms all partial combinations, validating that the synergy of all three expert types is essential to cover the full spectrum of anomalies.

Furthermore, Table~\ref{tab:esbir-ablation} confirms the importance of our auxiliary training modules. Removing the Expert Selection Balancing (ESB) module degrades performance by causing suboptimal router convergence and wasting model capacity. Similarly, removing the Expert Information Repulsion (EIR) module allows experts to learn redundant features, diminishing the benefits of specialization. Collectively, these ablations empirically confirm that each core component of AnomalyMoE, including its hierarchical structure, selection balancing, and information repulsion, is essential for its state-of-the-art, generalist performance.

\subsection{Qualitative Analysis and Visualization}
\begin{figure}[!h]
  \centering
   \includegraphics[width=0.95\linewidth]{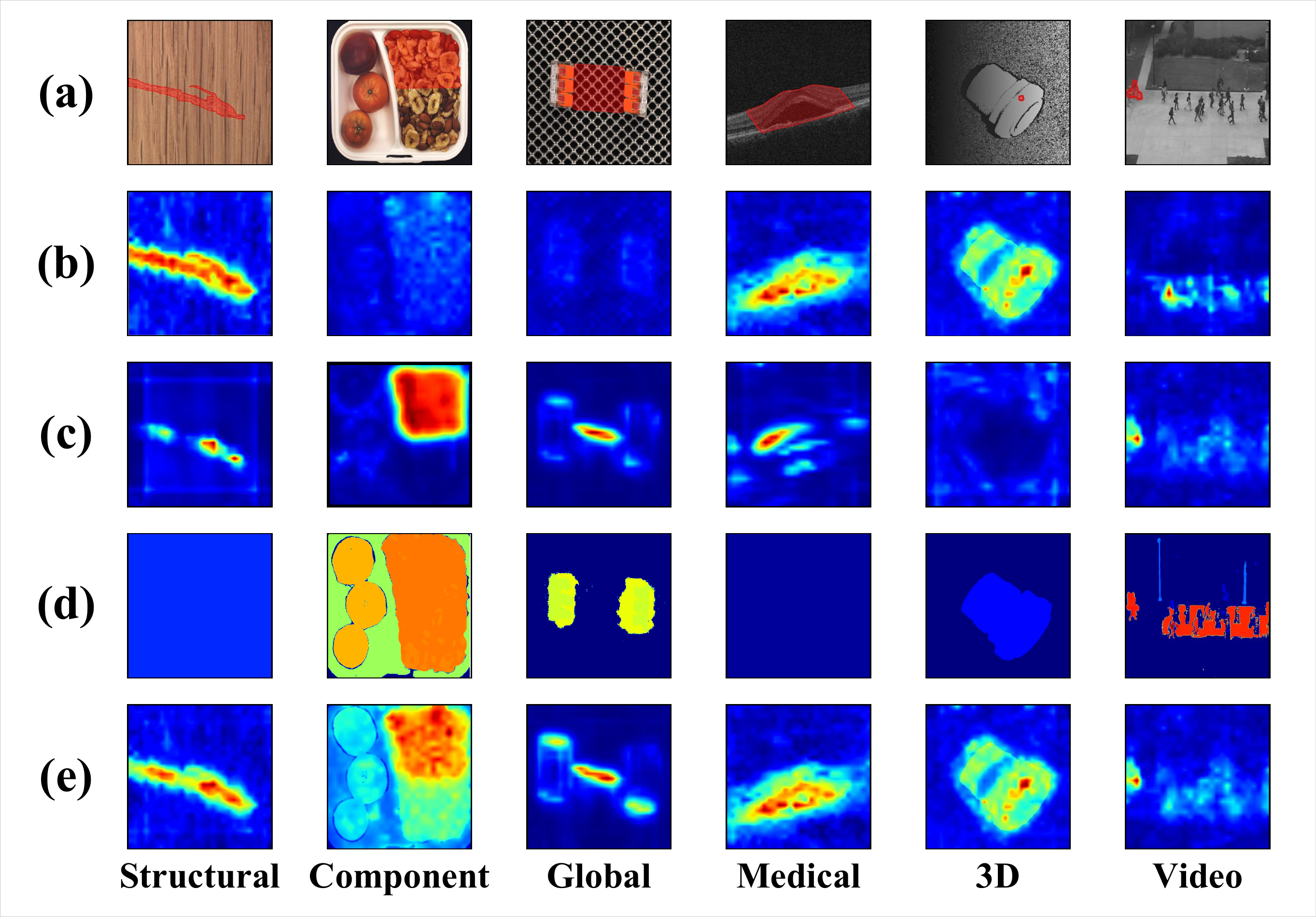}
   \caption{Visualization of the anomaly detection process. 
    Rows from top to bottom: (a) Input, (b) Patch-level expert output, (c) Global-level expert output, (d) Component-level expert output, and (e) final AnomalyMoE result.}
   \label{fig:visual}
\end{figure}

To provide an intuitive understanding of our framework, we conduct a qualitative analysis of the router's behavior and the experts' specialization. As shown in Figure~\ref{fig:router_analysis}, the router learns to dynamically allocate resources, assigning higher gate values to Patch-level experts on structurally anomalous datasets like MVTec AD, while prioritizing Global-level experts on logical datasets like MVTec-LOCO. Figure~\ref{fig:visual} provides direct visual confirmation of this functional specialization. For a structural defect (e.g., `wood'), the Patch-level expert generates a precise activation, whereas for a logical missing part anomaly (`splicing\_connectors'), the Global-level expert correctly identifies the compositional error. In all cases, the final aggregated result effectively integrates the most salient signals from the relevant experts. These visualizations clearly illustrate the complementary and synergistic nature of our hierarchical design.

\section{Conclusion}
We introduce AnomalyMoE, a novel, language-free generalist framework designed to address the fragmented nature of visual anomaly detection. By decomposing anomalies into a three-level semantic hierarchy, our Mixture-of-Experts architecture leverages specialized patch, component, and global experts to achieve comprehensive detection capabilities. Governed by a sophisticated routing mechanism enhanced by our proposed Expert Selection Balancing and Expert Information Repulsion modules, AnomalyMoE learns a diverse and functionally disentangled set of representations. Extensive experiments on eight challenging datasets demonstrate that our approach not only establishes a new state-of-the-art but also consistently outperforms specialized methods in their respective domains. AnomalyMoE represents a significant step towards a truly universal, efficient, and scalable anomaly detection system, paving the way for more robust real-world applications.

{
    \small
    \bibliographystyle{ieeenat_fullname}
    \bibliography{main}
}

\maketitlesupplementary

\setcounter{table}{0}
\setcounter{section}{0}
\setcounter{figure}{0}
\renewcommand\thesection{\Alph{section}}
\renewcommand{\thetable}{A\arabic{table}}
\renewcommand{\thefigure}{A\arabic{figure}}

\section{Detail of CLUB Estimates}

Our Expert Information Repulsion (EIR) module is designed to promote functional specialization among experts by minimizing the Mutual Information (MI) between their output representations. Here we provides the theoretical background for the Contrastive Log-ratio Upper Bound (CLUB) estimator, which forms the basis of our EIR loss.

\noindent\textbf{Theoretical Foundation of CLUB. }
Given two random variables, $Z_j$ and $Z_k$, representing the outputs of two different experts, their mutual information $I(Z_j; Z_k)$ quantifies their statistical dependency. It is defined as:
\begin{equation}
    I(Z_j; Z_k) = \mathbb{E}_{p(z_j, z_k)} \left[ \log \frac{p(z_k | z_j)}{p(z_k)} \right],
\end{equation}
where $p(z_j, z_k)$ is the joint probability distribution. Minimizing this value encourages the variables to become independent. However, direct computation of MI is generally intractable as it requires knowledge of the marginal distribution $p(z_k)$.

The CLUB estimator~\cite{cheng2020club} provides a tractable upper bound on MI by introducing a variational approximation network, $q_{\theta}(z_k | z_j)$, to model the true conditional probability $p(z_k | z_j)$. The CLUB upper bound is derived as follows:
\begin{equation}
\label{eq:club_main}
\begin{split}
    I(Z_j; Z_k) &\le \mathbb{E}_{p(z_j, z_k)}[\log q_{\theta}(z_k | z_j)] \\
    &\quad - \mathbb{E}_{p(z_j)p(z_k)}[\log q_{\theta}(z_k | z_j)] \\
    &\triangleq I_{\text{CLUB}}(Z_j; Z_k).
\end{split}
\end{equation}
This formulation has an intuitive, contrastive interpretation: the first term maximizes the log-likelihood of ``positive'' pairs drawn from the joint distribution, while the second term minimizes the log-likelihood of ``negative'' pairs drawn from the product of marginals. By minimizing this upper bound during training, we effectively minimize the true mutual information between the expert representations.

\noindent\textbf{Implementation of the CLUB Estimator. }
Our implementation of the EIR loss, $\mathcal{L}_{\text{EIR}}$, is a direct application of the CLUB estimate. We design a ``CLUBSample'' module, our variational network $q_\theta$, to approximate the conditional distribution. This module consists of two separate Multi-Layer Perceptrons~(MLPs): one that predicts the mean $\mu_\theta(z_j)$ and another that predicts the log-variance $\log \sigma^2_\theta(z_j)$ of a Gaussian distribution, given an input feature vector $z_j$. The log-likelihood $\log q_{\theta}(z_k | z_j)$ can then be calculated using the Gaussian probability density function.

The final EIR loss for a pair of expert outputs is then computed via Monte Carlo estimation over a batch of samples, as shown in Equation~\ref{eq:club_main}. The ``negative" pairs for the second term are constructed by randomly permuting the batch dimension of $z_k$, which effectively samples from the marginal distribution. The total $\mathcal{L}_{\text{EIR}}$ used in our final training objective is the sum of these pairwise estimates across all unique pairs of experts within each of the three hierarchical groups.

\section{Dataset Details}
This section provides a detailed description of the eight public datasets used in our experiments to evaluate the performance and generalizability of AnomalyMoE.

\noindent\textbf{MVTec AD. }
The MVTec AD~\cite{bergmann2021mvtec} dataset is one of the most popular benchmarks for industrial anomaly detection. It consists of 5,354 high-resolution images across 15 different object and texture categories, such as leather, wood, metal components, and pills. The dataset is split into a training set containing only normal images and a test set containing both normal and anomalous images, with resolutions ranging from 700$\times$700 to 1024$\times$1024 pixels.

\noindent\textbf{VisA. }
The VisA~\cite{zou2022spot} dataset is a more challenging real-world industrial anomaly detection dataset, containing 10,821 images across 12 object categories. It includes 9,621 normal and 1,200 anomalous samples with complex structures and multiple instances per image. The anomalies are often small and subtle, making it a robust testbed for evaluating a model's generalization capabilities. The image resolutions are approximately 1500$\times$1000 pixels.

\noindent\textbf{MVTec 3D-AD. }
The MVTec 3D-AD~\cite{bergmann2021mvtec3d} dataset consists of 10 real-world categories with over 4,000 instances. Each instance includes both high-resolution RGB images and corresponding registered 3D point cloud data. The anomalies are diverse, ranging from surface defects to structural variations.

\noindent\textbf{MVTec-LOCO. }
The MVTec-LOCO~\cite{bergmann2022beyond} dataset is a dataset designed specifically for logical anomaly detection. It is composed of 1,772 normal and 369 anomalous images across 5 object categories. Unlike traditional datasets, anomalies in MVTec-LOCO are defined by the violation of logical constraints, such as the addition, omission, or incorrect combination of elements, while the individual components themselves remain structurally perfect. The dataset provides pixel-level annotations for these logical anomalies.

\noindent\textbf{BrainMRI. }
The BrainMRI dataset~\cite{baid2021rsna} is derived from the BraTS 2021 challenge, a large-scale brain tumor segmentation benchmark. It contains 2D slices from 3D brain volume MRI scans. The training set includes 7,500 normal samples, and the test set contains 3,715 samples, both normal and anomalous, with each slice measuring 240$\times$240 pixels. The dataset provides pixel-level annotations for the tumor regions, which are treated as anomalies.

\begin{table*}[t]
\centering
\caption{Per-category performance comparison on the MVTec AD dataset. All values are AUC scores in percent (\%). For each category, the best performance is highlighted in \textbf{bold} and the second best is \underline{underlined}.}
\label{tab:mvtec_ad_detailed}
\begin{tabular}{@{}llccccccc@{}}
\toprule
Metric & Category & UniAD & Recontrast & UniNet & Dinomaly & UniVAD & LogSAD & AnomalyMoE \\ 
\midrule
\multirow{16}{*}{\begin{tabular}[c]{@{}c@{}}Image-level\\ (AUC)\end{tabular}} & Carpet & \textbf{100.0} & 96.9 & 99.4 & 99.9 & \textbf{100.0} & 99.4 & \underline{99.7} \\
& Grid & 95.3 & 90.9 & 99.1 & \underline{99.7} & \textbf{100.0} & \textbf{100.0} & 98.2 \\
& Leather & \textbf{100.0} & \textbf{100.0} & 88.2 & \textbf{100.0} & \textbf{100.0} & \textbf{100.0} & \underline{99.7} \\
& Tile & 96.5 & \underline{99.9} & 88.5 & \textbf{100.0} & \underline{99.9} & 99.1 & \textbf{100.0} \\
& Wood & 97.3 & \underline{99.8} & 89.9 & \textbf{99.9} & \underline{99.8} & 99.5 & \textbf{99.9} \\
& Bottle & 99.5 & \textbf{100.0} & 94.2 & \textbf{100.0} & \underline{99.8} & \textbf{100.0} & \textbf{100.0} \\
& Cable & 78.1 & 87.3 & 78.6 & \underline{99.2} & 94.4 & 95.6 & \textbf{99.8} \\
& Capsule & 67.9 & 85.7 & 69.3 & 95.0 & \underline{97.4} & 94.1 & \textbf{97.6} \\
& Hazelnut & 96.4 & \textbf{100.0} & 99.3 & 99.8 & 99.4 & \underline{99.9} & \textbf{100.0} \\
& Metal Nut & 91.9 & 99.3 & 91.2 & \textbf{100.0} & 99.1 & \textbf{100.0} & \underline{99.9} \\
& Pill & 70.7 & 92.7 & 74.0 & \underline{98.5} & 97.0 & 87.0 & \textbf{99.1} \\
& Screw & 52.3 & 86.9 & 58.1 & \underline{93.0} & 85.4 & 83.3 & \textbf{99.0} \\
& Toothbrush & 91.4 & 94.2 & 81.1 & \textbf{100.0} & \underline{99.2} & 92.2 & \textbf{100.0} \\
& Transistor & 86.1 & 92.4 & 69.2 & \underline{97.3} & 95.5 & 87.6 & \textbf{99.0} \\
& Zipper & 89.7 & 98.4 & 88.1 & \textbf{100.0} & \underline{99.9} & 97.6 & \textbf{100.0} \\ 
\midrule
& \textit{Mean} & 87.5 & 95.0 & 84.7 & \underline{98.8} & 97.8 & 96.9 & \textbf{99.5} \\
\midrule
\multirow{16}{*}{\begin{tabular}[c]{@{}c@{}}Pixel-level\\ (AUC)\end{tabular}} & Carpet & 98.6 & 98.5 & 95.5 & \textbf{99.4} & \underline{99.2} & 99.0 & \textbf{99.4} \\
& Grid & 94.1 & 97.6 & 90.0 & \textbf{99.3} & \textbf{99.3} & \underline{99.2} & 98.9 \\
& Leather & 99.0 & \textbf{99.4} & 83.7 & 99.2 & 99.1 & \underline{99.3} & 98.8 \\
& Tile & 89.0 & 94.9 & 90.4 & \textbf{98.2} & 96.6 & \underline{97.1} & 96.3 \\
& Wood & 92.8 & 95.4 & 85.6 & \textbf{97.7} & \underline{97.3} & 94.5 & 96.3 \\
& Bottle & 97.7 & 97.1 & 91.0 & \underline{98.9} & 97.3 & \textbf{99.0} & \underline{98.9} \\
& Cable & 92.2 & 90.4 & 88.3 & 94.1 & 95.7 & \underline{97.3} & \textbf{98.1} \\
& Capsule & 96.6 & \underline{98.3} & 88.1 & 98.2 & 96.8 & 98.2 & \textbf{98.4} \\
& Hazelnut & 96.3 & 98.8 & 97.4 & \underline{99.2} & \textbf{99.4} & \textbf{99.4} & 98.7 \\
& Metal Nut & 86.9 & 95.4 & 93.0 & 96.3 & 94.4 & \textbf{97.7} & \underline{97.4} \\
& Pill & 88.3 & \textbf{98.0} & 96.2 & \underline{97.1} & 94.6 & 95.0 & 96.5 \\
& Screw & 93.9 & 98.3 & 96.0 & \underline{99.2} & 98.6 & 95.2 & \textbf{99.6} \\
& Toothbrush & 97.2 & 98.4 & 93.0 & \textbf{99.0} & \underline{98.8} & 95.9 & 98.0 \\
& Transistor & \textbf{95.7} & 86.5 & 88.6 & 89.1 & 84.4 & \underline{94.0} & 92.5 \\
& Zipper & 93.9 & 97.2 & 90.5 & \textbf{99.1} & 95.5 & 96.7 & \underline{98.5} \\
\midrule
& \textit{Mean} & 94.2 & 96.3 & 91.2 & \underline{97.6} & 96.5 & 97.0 & \textbf{97.7} \\
\bottomrule
\end{tabular}
\end{table*}

\begin{table*}[t]
\centering
\caption{Per-category performance comparison on the VisA dataset. All values are AUC scores in percent (\%). For each category, the best performance is highlighted in \textbf{bold} and the second best is \underline{underlined}.}
\label{tab:visa_detailed}
\begin{tabular}{@{}llccccccc@{}}
\toprule
Metric & Category & UniAD & Recontrast & UniNet & Dinomaly & UniVAD & LogSAD & \textbf{AnomalyMoE} \\ 
\midrule
\multirow{13}{*}{\begin{tabular}[c]{@{}c@{}}Image-level\\ (AUC)\end{tabular}} & Candle & 87.4 & 90.4 & 79.2 & 96.4 & \underline{96.6} & 93.8 & \textbf{97.9} \\
& Capsules & 59.1 & 68.4 & 81.1 & 97.3 & \textbf{98.7} & 93.0 & \underline{97.4} \\
& Cashew & 85.6 & 89.1 & 88.8 & 90.8 & \underline{93.9} & 92.8 & \textbf{97.7} \\
& Chewinggum & 96.5 & 91.1 & 91.0 & \underline{98.0} & 99.5 & 98.5 & \textbf{99.8} \\
& Fryum & 79.0 & 93.9 & 88.4 & \underline{98.2} & 97.9 & 97.0 & \textbf{98.3} \\
& Macaroni1 & 73.0 & 91.7 & 90.6 & \underline{95.8} & 91.2 & 93.4 & \textbf{97.5} \\
& Macaroni2 & 62.0 & 76.1 & 78.3 & \underline{89.4} & 81.1 & 91.2 & \textbf{95.3} \\
& PCB1 & 86.3 & 96.1 & 82.5 & \textbf{98.9} & 97.1 & 89.8 & \underline{98.8} \\
& PCB2 & 78.0 & 95.2 & 89.0 & \textbf{98.7} & 88.9 & 87.3 & \underline{97.5} \\
& PCB3 & 70.6 & 94.6 & 92.8 & \underline{98.3} & 87.2 & 93.5 & \textbf{98.4} \\
& PCB4 & 95.1 & 97.9 & 99.4 & \underline{99.6} & 91.7 & 94.7 & \textbf{99.9} \\
& Pipe Fryum & 90.1 & 97.6 & 88.8 & 96.0 & \textbf{98.9} & 98.5 & \underline{98.7} \\ 
\midrule
& \textit{Mean} & 80.2 & 90.1 & 87.5 & \underline{95.7} & 93.5 & 94.5 & \textbf{98.1} \\
\midrule
\multirow{13}{*}{\begin{tabular}[c]{@{}c@{}}Pixel-level\\ (AUC)\end{tabular}} & Candle & 98.2 & 98.2 & 95.5 & 99.2 & \underline{99.4} & 98.5 & \textbf{99.5} \\
& Capsules & 88.2 & 98.0 & 97.1 & \underline{99.5} & 97.9 & 96.9 & \textbf{99.7} \\
& Cashew & 97.7 & 70.3 & 38.2 & 84.5 & \underline{98.7} & 97.6 & \textbf{99.3} \\
& Chewinggum & 98.9 & 76.3 & 67.8 & 98.8 & \textbf{99.7} & \underline{99.5} & 99.4 \\
& Fryum & 94.8 & 87.2 & 93.4 & 95.8 & \textbf{97.2} & 92.6 & \underline{97.0} \\
& Macaroni1 & 97.3 & 92.2 & 99.2 & \underline{99.5} & \textbf{99.7} & 97.6 & \textbf{99.7} \\
& Macaroni2 & 93.5 & 94.0 & 98.8 & \underline{99.5} & 98.9 & 96.2 & \textbf{99.8} \\
& PCB1 & 98.8 & 94.1 & 92.5 & \underline{99.3} & 99.0 & 97.1 & \textbf{99.5} \\
& PCB2 & 96.5 & \underline{97.7} & 94.3 & 97.0 & 97.0 & 97.6 & \textbf{98.2} \\
& PCB3 & 96.5 & 93.3 & 93.6 & 97.3 & 96.0 & \underline{97.9} & \textbf{98.3} \\
& PCB4 & 95.7 & 95.4 & 90.4 & \underline{97.9} & 96.8 & 95.5 & \textbf{98.1} \\
& Pipe Fryum & 98.9 & 97.1 & 96.1 & 98.8 & 97.9 & \underline{99.0} & \textbf{99.2} \\
\midrule
& \textit{Mean} & 96.3 & 90.8 & 87.3 & 96.4 & \underline{98.2} & 97.1 & \textbf{99.0} \\
\bottomrule
\end{tabular}
\end{table*}

\begin{table}[!h]
\centering
\caption{Inference time (in milliseconds per image) comparison between AnomalyMoE and recent language-model-based unified anomaly detection methods. All experiments are conducted on a single NVIDIA A6000 GPU.}
\label{tab:inference_speed}
\begin{tabular}{@{}lc@{}}
\toprule
Method                       & Inference Time (ms) \\ \midrule
UniVAD~\cite{gu2025univad}      & 658.5               \\
LogSAD~\cite{zhang2025towards} & 1932.6              \\
\textbf{AnomalyMoE (ours)}   & \textbf{58.5}       \\ \bottomrule
\end{tabular}
\end{table}

\begin{table*}[t]
\centering
\caption{Per-category performance comparison on the MVTec-LOCO dataset. All values are AUC scores in percent (\%). For each category, the best performance is highlighted in \textbf{bold} and the second best is \underline{underlined}.}
\label{tab:mvtec_loco_detailed}
\begin{tabular}{@{}llccccccc@{}}
\toprule
Metric & Category & UniAD & Recontrast & UniNet & Dinomaly & UniVAD & LogSAD & \textbf{AnomalyMoE} \\ 
\midrule
\multirow{6}{*}{\begin{tabular}[c]{@{}c@{}}Image-level\\ (AUC)\end{tabular}} & Breakfast Box & 76.7 & 72.5 & 73.2 & 86.0 & 84.1 & \underline{87.9} & \textbf{91.1} \\
& Juice Bottle & 89.8 & 93.6 & \textbf{95.8} & 87.2 & 62.9 & 91.9 & \underline{94.7} \\
& Pushpins & 58.2 & 76.1 & 73.2 & 66.8 & 66.8 & \underline{83.8} & \textbf{88.7} \\
& Screw Bag & 51.3 & \underline{71.0} & 67.5 & 69.3 & 61.3 & \textbf{76.5} & 69.9 \\
& Splicing Connectors & 80.7 & 82.1 & 79.4 & 81.5 & 80.0 & \underline{90.9} & \textbf{92.9} \\ 
& \textit{Mean} & 71.3 & 79.1 & 77.8 & 78.2 & 71.0 & \underline{86.2} & \textbf{87.5} \\
\midrule
\multirow{6}{*}{\begin{tabular}[c]{@{}c@{}}Pixel-level\\ (AUC)\end{tabular}} & Breakfast Box & \underline{89.6} & 84.4 & 83.9 & 85.1 & 85.8 & \underline{89.6} & \textbf{89.9} \\
& Juice Bottle & \textbf{91.6} & 90.2 & \underline{90.9} & 86.5 & 84.7 & 88.0 & \textbf{91.6} \\
& Pushpins & 54.0 & 59.4 & 60.3 & 65.2 & 71.5 & \underline{72.4} & \textbf{84.9} \\
& Screw Bag & 68.2 & 69.2 & 66.5 & 57.9 & 64.6 & \underline{71.4} & \textbf{77.9} \\
& Splicing Connectors & \underline{78.6} & 65.3 & 64.2 & 64.8 & 69.1 & 75.5 & \textbf{86.9} \\
& \textit{Mean} & 76.4 & 74.1 & 73.2 & 71.5 & 75.1 & \underline{79.4} & \textbf{86.2} \\
\bottomrule
\end{tabular}
\end{table*}

\noindent\textbf{LiverCT. }
The LiverCT dataset~\cite{bilic2023liver} is constructed from the BTCV and LiTS datasets and contains abdominal 3D CT scans. Hounsfield Unit (HU) values are converted to grayscale using an abdominal window, and the volumes are cropped into 2D slices. The dataset includes 1,452 normal 2D slices for training and 1,493 2D slices for testing, which include both normal and anomalous samples. The images have a resolution of 512$\times$512 pixels and are provided with pixel-level anomaly annotations for liver lesions.

\noindent\textbf{RESC. }
The Retinal Edema Segmentation Challenge (RESC)~\cite{hu2019automated} dataset is a retinal Optical Coherence Tomography dataset. It contains 4,297 normal images for training and 1,805 test images that include normal and anomalous samples. The anomalies correspond to retinal pathologies. The images have a resolution of 512$\times$1024 pixels, and the dataset provides pixel-level anomaly annotations.

\noindent\textbf{UCSD Ped2. }
The UCSD Ped2~\cite{wang2010anomaly} dataset is a widely used benchmark for video anomaly detection in crowded scenes. It consists of 16 training and 12 testing video clips, totaling thousands of frames. The training data contains only normal footage of pedestrians walking, while the test data includes anomalies such as cyclists, skaters, and small carts.

\section{Complexity and Inference Speed Analysis}

To demonstrate the practical advantages of our language-free design, we conduct an analysis of the inference speed of AnomalyMoE compared to unified methods that rely on large vision-language or language models. The comparison focuses on the time required to process a single image.

The results, presented in Table~\ref{tab:inference_speed}, highlight a significant efficiency gap. AnomalyMoE achieves a processing time of just \textbf{58.5 ms} per image. In contrast, UniVAD~\cite{gu2025univad} requires 658.5 ms, making our method approximately 11.3 times faster. The difference is even more pronounced when compared to LogSAD~\cite{zhang2025towards}, which has an inference time of 1932.6 ms; AnomalyMoE is about 33.0 times faster.

This substantial speed-up is a direct result of our architectural choices. By relying exclusively on efficient, vision-native modules and avoiding the significant computational overhead associated with large language models and complex cross-modal interactions, AnomalyMoE delivers state-of-the-art accuracy without sacrificing deployment practicality. This makes our framework particularly well-suited for real-world scenarios where both high performance and low latency are critical requirements.

\begin{figure*}[!h]
  \centering
   \includegraphics[width=0.95\linewidth]{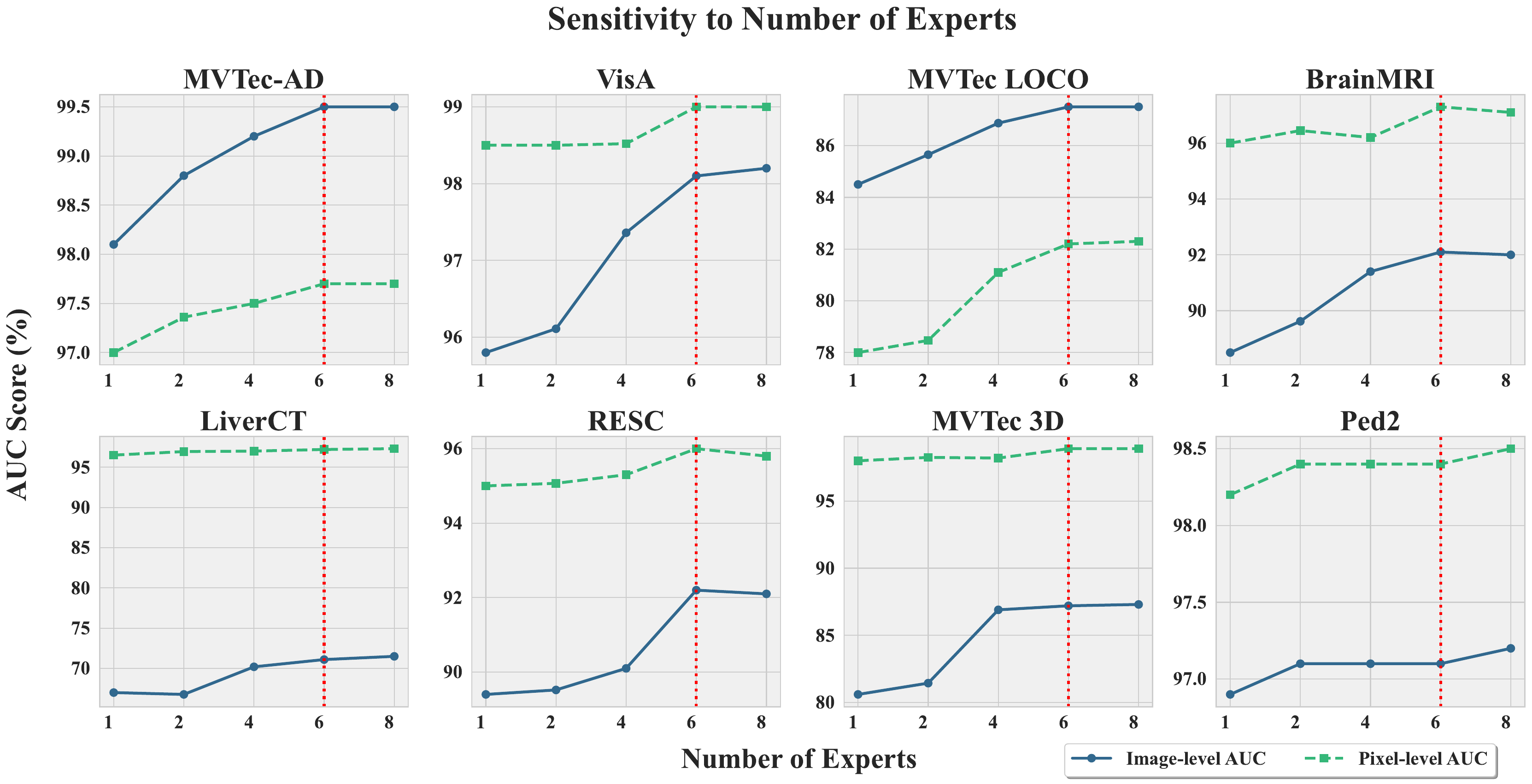}
   \caption{Sensitivity analysis of the number of experts per group ($N_{exp}$). We evaluate performance across all eight datasets as $N_{exp}$ is varied. The configuration $N_{exp}=6$ (our final choice) is highlighted with a vertical dashed line.}
   \label{fig:expert_counts}
\end{figure*}

\begin{figure*}[!h]
  \centering
   \includegraphics[width=\linewidth]{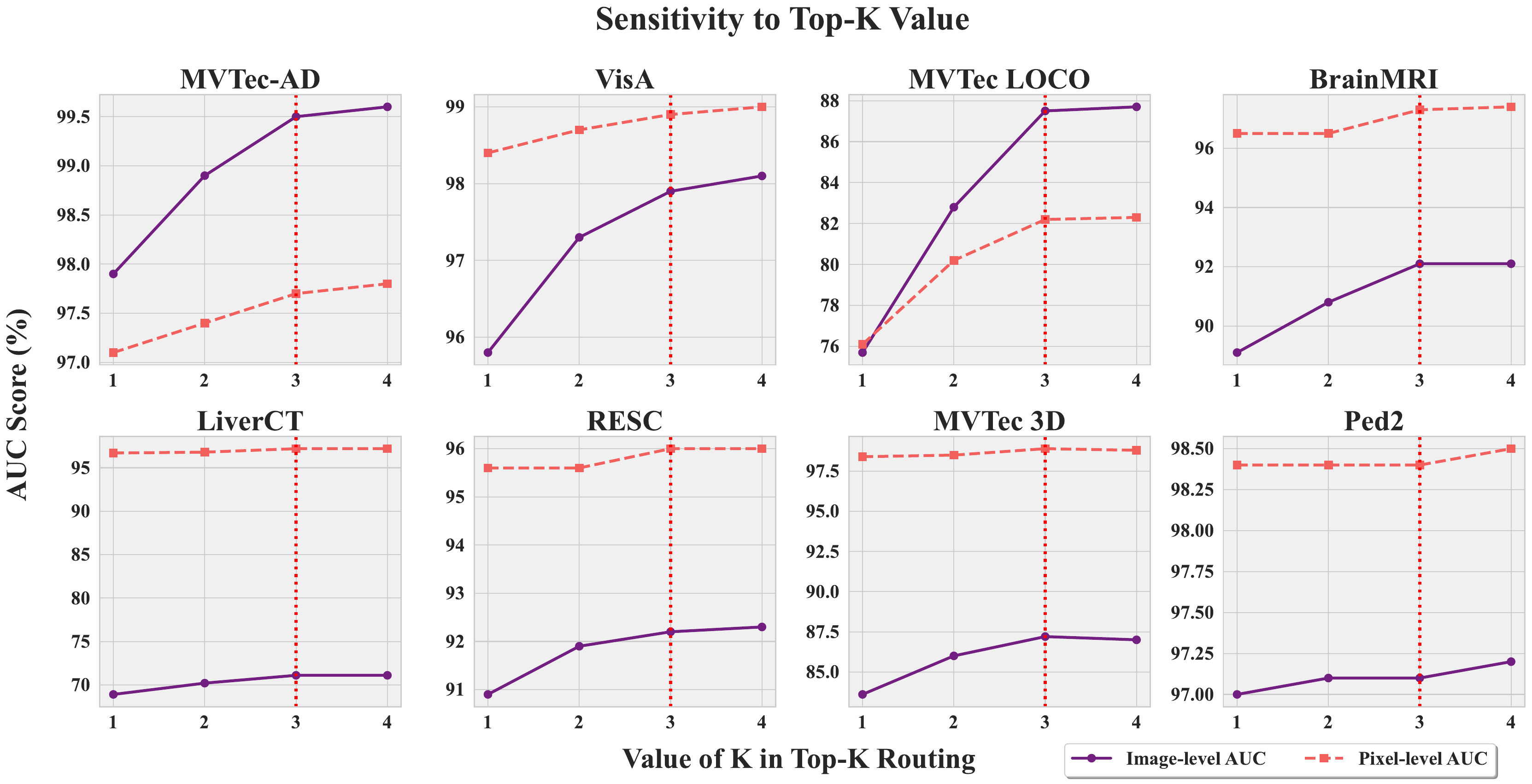}
   \caption{Sensitivity analysis of the Top-K value. Performance is evaluated as $K$ is varied from 1 to 4. Our chosen value, $K=3$, which allows for one expert from each hierarchical level to be activated, is highlighted.}
   \label{fig:top_k}
\end{figure*}

\begin{figure*}[!h]
  \centering
   \includegraphics[width=\linewidth]{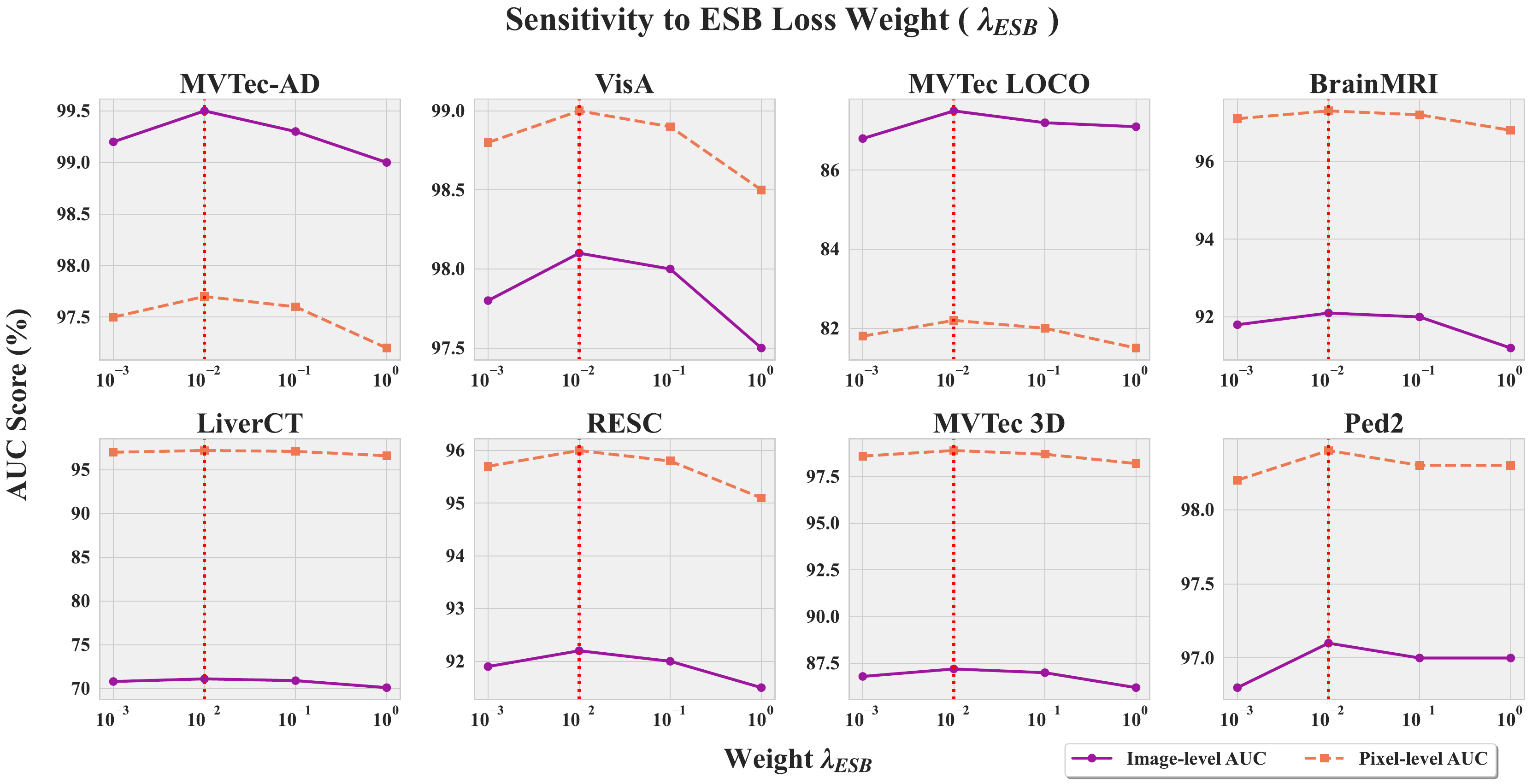}
   \caption{Sensitivity analysis of the ESB loss weight, $\lambda_{\text{ESB}}$. The model's performance is stable across a range of values, with the optimal choice, $\lambda_{\text{ESB}}=0.01$, highlighted.}
   \label{fig:lambda_esb}
\end{figure*}

\begin{figure*}[!h]
  \centering
   \includegraphics[width=\linewidth]{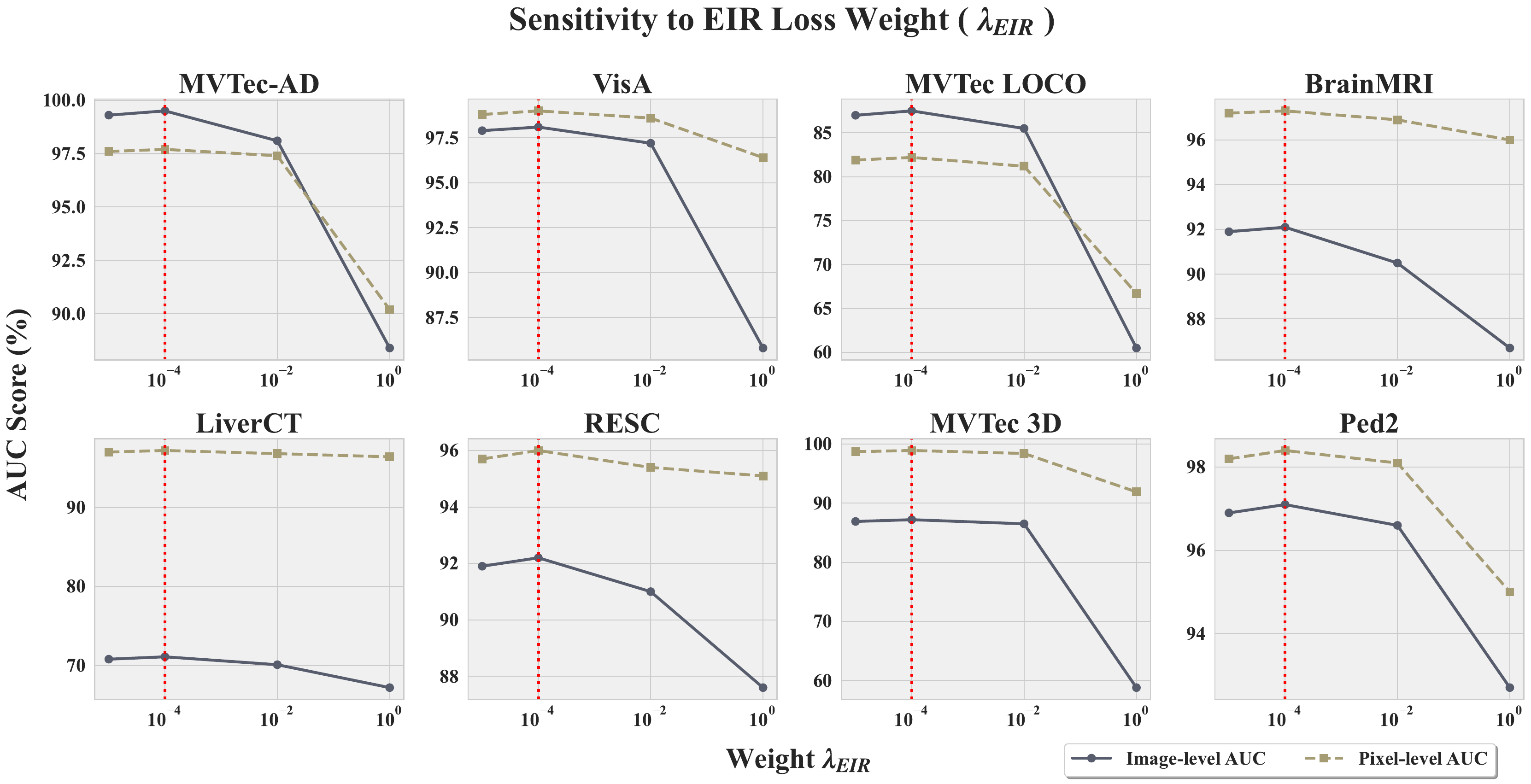}
   \caption{Sensitivity analysis of the EIR loss weight, $\lambda_{\text{EIR}}$. The results show that the model is sensitive to this hyperparameter, with a large value causing a significant drop in performance. The optimal value, $\lambda_{\text{EIR}}=0.0001$, is highlighted.}
   \label{fig:lambda_eir}
\end{figure*}

\begin{figure*}[!h]
\centering
   \includegraphics[width=\linewidth]{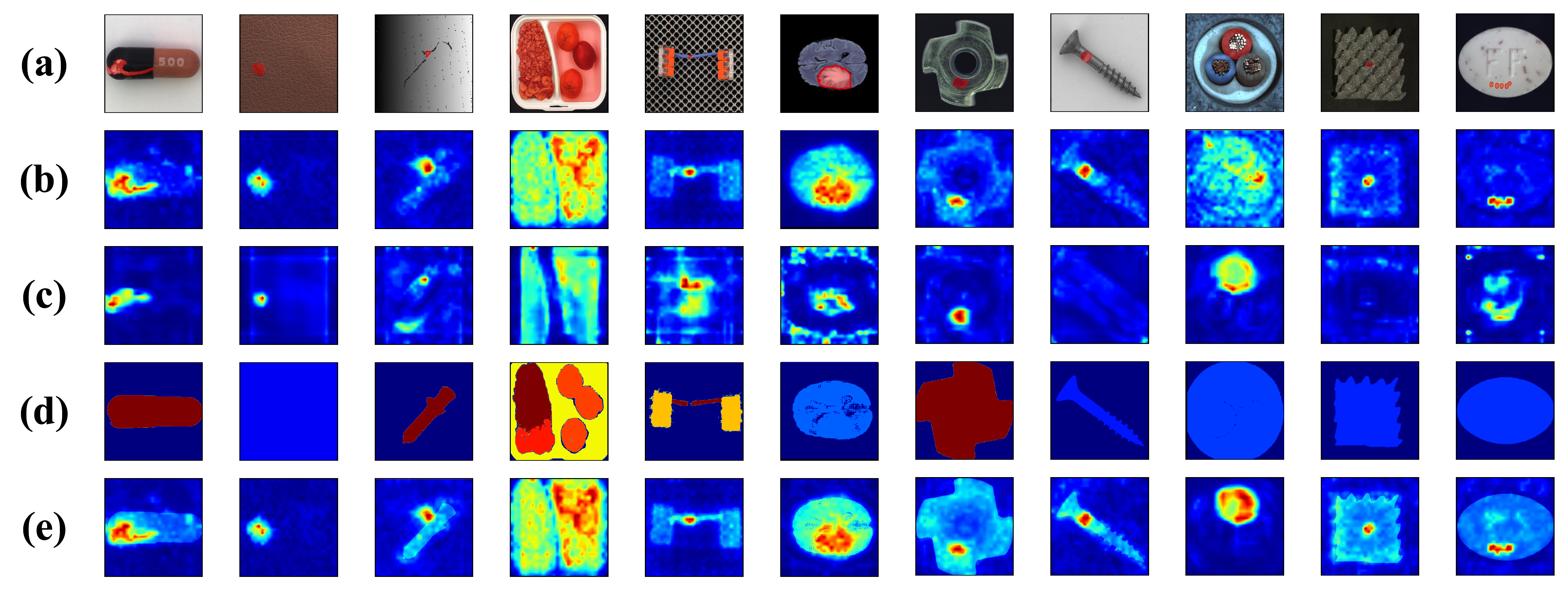} 
   \caption{
        Additional qualitative visualizations of AnomalyMoE across diverse datasets.
        For each sample, we display (from top to bottom): (a) Input, (b) Patch-level expert output, (c) Global-level expert output, (d) Component-level expert output, and (e) final AnomalyMoE result.
    }
   \label{fig:additional_qualitative}
\end{figure*}

\section{More Experiment Results}
\subsection{Detailed Per-Category Performance}

In this section, we provide a detailed per-category breakdown of our model's performance on the MVTec-AD~\cite{bergmann2021mvtec}, VisA~\cite{zou2022spot}, and MVTec-LOCO~\cite{bergmann2022beyond} datasets. While the main paper presents mean performance, this granular analysis is crucial for demonstrating the consistency and robustness of AnomalyMoE, ensuring that strong average scores are not masking weaknesses on specific categories. These three datasets are selected as they represent distinct and complementary challenges: MVTec AD for structural defects, VisA for complex real-world scenes, and MVTec-LOCO for high-level logical anomalies.

\noindent\textbf{MVTec AD. }
As shown in Table~\ref{tab:mvtec_ad_detailed}, AnomalyMoE demonstrates outstanding and consistent performance across the 15 categories of MVTec AD. Notably, on challenging object categories with subtle defects like `Screw', `Pill', and `Transistor', our model achieves the highest image-level AUC, surpassing even highly specialized reconstruction methods like Dinomaly. This is a direct testament to the superiority of our hierarchical MoE architecture. The collaboration between the Patch-level expert, which captures fine-grained texture deviations, and the Component-level expert, which understands part integrity, allows AnomalyMoE to build a more comprehensive model of normality. On texture-based categories like `Grid', Dinomaly's singular focus on patch reconstruction gives it a slight edge. However, our model's consistently strong performance across both object and texture classes underscores its greater versatility.

\noindent\textbf{VisA. }
The results on the more challenging VisA dataset, presented in Table~\ref{tab:visa_detailed}, further highlight the robustness of AnomalyMoE. Our model achieves the best image-level and pixel-level average AUC, outperforming all baselines. On categories with both structural and semantic variations, such as `Candle' and `Cashew', AnomalyMoE shows a clear advantage. This is because our framework does not rely on a single detection paradigm. The dynamic router can allocate more weight to the Component and Global experts when semantic understanding is required, a flexibility that monolithic reconstruction models lack. While some language-based models like UniVAD perform well on certain categories (e.g., `Capsules'), their performance is less consistent, demonstrating the advantage of our language-free, multi-level visual analysis.

\noindent\textbf{MVTec-LOCO. }
The MVTec-LOCO dataset provides the ultimate test for logical anomaly detection, and the results in Table~\ref{tab:mvtec_loco_detailed} are particularly revealing. AnomalyMoE achieves the highest mean AUC in both image-level and pixel-level tasks, even outperforming LogSAD, a method specifically designed for this type of problem. On categories that require a holistic understanding of spatial relationships, such as `Pushpins' (correct number and placement) and `Splicing Connectors' (correct assembly), our Global-level expert proves indispensable, leading to a significant performance lead. For the `Screw Bag' category, the language-based LogSAD holds a performance advantage, likely due to its ability to leverage explicit, text-based priors about the expected number of screws. Our model, being purely visual, relies on learning these patterns implicitly. Nonetheless, AnomalyMoE's exceptional overall performance on this benchmark validates the power of our hierarchical decomposition, proving that a well-designed, multi-level visual system can effectively solve complex logical anomalies without external knowledge.

\subsection{Hyperparameter Sensitivity Analysis}

We conduct a series of experiments to analyze the sensitivity of AnomalyMoE to its key hyperparameters and to justify our final configuration choices. These analyses cover the number of experts per hierarchical level, the Top-K routing value, and the weights for our auxiliary ESB and EIR losses.

\noindent\textbf{Number of Experts per Group. }
We first investigate the impact of the total number of experts in each group, denoted as $N_{exp}$ (where we set $N_p=N_{exp}, N_c=N_{exp}, N_g=N_{exp}$). As shown in Figure~\ref{fig:expert_counts}, performance generally improves as we increase the number of experts from 1 to 6. This suggests that a larger pool of experts allows the model to learn more diverse and specialized representations. However, increasing the number of experts from 6 to 8 yields only marginal gains. Therefore, we select $N_{exp}=6$ as our final configuration, as it provides an optimal balance between model capacity and performance.

\noindent\textbf{Value of K in Top-K Routing. }
The choice of $K$, the number of activated experts per sample, is critical to our hierarchical design. The results, visualized in Figure~\ref{fig:top_k}, show that performance is suboptimal for $K < 3$. When $K=1$ or $K=2$, the model is forced to choose only a subset of its hierarchical capabilities for each sample, which is particularly detrimental on datasets like MVTec-LOCO that require a simultaneous understanding of local and global context. By setting $K=3$, we enable the router to potentially activate one expert from each of the three distinct groups. While increasing $K$ to 4 provides a minor improvement on some datasets, we select $K=3$ as it represents the most architecturally coherent and computationally efficient choice.

\noindent\textbf{Weights for ESB and EIR Losses. }
The auxiliary loss weights, $\lambda_{\text{ESB}}$ and $\lambda_{\text{EIR}}$, control the balance between the primary reconstruction task and MoE regularization. The results in Figure~\ref{fig:lambda_esb} and Figure~\ref{fig:lambda_eir} demonstrate a clear trade-off. For the ESB loss, a very small weight ($\lambda_{\text{ESB}}=0.001$) is insufficient to prevent router imbalance, while a large weight ($\lambda_{\text{ESB}}=1.0$) over-penalizes the router, distracting it from the main objective. A value of $\lambda_{\text{ESB}}=0.01$ provides the best balance. A similar, but more sensitive, dynamic is observed for the EIR loss. A large weight ($\lambda_{\text{EIR}}=1.0$) proves highly destructive, causing a catastrophic performance collapse by forcing the model to prioritize feature repulsion over reconstruction. Optimal performance is achieved at a small value of $\lambda_{\text{EIR}}=0.0001$. This highlights the delicate but crucial role these auxiliary losses play in stabilizing and guiding the MoE training process.

\subsection{Additional Qualitative Visualizations}
To further demonstrate the generalist capabilities and functional specialization of our hierarchical experts, we provide additional qualitative results in Figure~\ref{fig:additional_qualitative}. This figure showcases the performance of AnomalyMoE on a wider variety of samples drawn from the diverse datasets.

\end{document}